\documentclass[runningheads]{llncs}
\usepackage[T1]{fontenc}
\usepackage{graphicx}
\usepackage[utf8]{inputenc} 
\usepackage[T1]{fontenc}    
\usepackage{hyperref}       
\usepackage{url}            
\usepackage{booktabs}       
\usepackage{amsfonts}       
\usepackage{nicefrac}       
\usepackage{microtype}      
\usepackage{xcolor}         

\usepackage{amsmath}
\usepackage{algpseudocode}

\usepackage{graphicx}

\usepackage{adjustbox}
\usepackage{subfig}

\usepackage{dblfloatfix} 

\usepackage[font={footnotesize}]{caption}

\usepackage[ruled,vlined]{algorithm2e}
\usepackage{wrapfig}

\usepackage{tikz}

\usepackage[citestyle=numeric]{biblatex}
\bibliography{1_refs}

\begin{document}
\title{The Hidden Influence of Latent Feature Magnitude When Learning with Imbalanced Data}
\titlerunning{Hidden Influence of Latent Feature Magnitude}
%
\author{Damien A. Dablain\inst{1}
\and
Nitesh V. Chawla\inst{1}
}
\authorrunning{Dablain \& Chawla}
%
\institute{Dept. Computer Science   \& Lucy Family Institute for Data and Society, University of Notre Dame, Notre Dame, IN 46553 
\email{ddablain and nchawla@nd.edu}\\
}
\maketitle              
\begin{abstract}
Machine learning (ML) models have difficulty \textit{generalizing} when the number of training class instances are numerically imbalanced. The problem of generalization in the face of data imbalance has largely been attributed to the lack of training data for under-represented classes and to feature overlap. The typical remedy is to implement data augmentation for classes with fewer instances or to assign a higher cost to minority class prediction errors or to undersample the prevalent class. However, we show that one of the central causes of impaired generalization when learning with imbalanced data is the \textit{inherent} manner in which ML models perform inference. These models have difficulty generalizing due to their heavy reliance on the \textit{magnitude} of encoded signals. During inference, the models predict classes based on a combination of encoded signal \textit{magnitudes} that linearly sum to the largest scalar. 
We demonstrate that even with aggressive data augmentation, which generally improves minority class prediction accuracy, parametric ML models \textit{still} associate a class label with a limited number of feature combinations that sum to a prediction, which can affect generalization.

\keywords{Machine Learning  \and Imbalanced Data \and Deep Learning}
\end{abstract}
%
%
%

\section{Introduction}
Biological systems transmit neural signals through  \textit{spikes}, where  \textit{timing} is a key component of the selection and movement of signals through the network \cite{abeles1993spatiotemporal,sejnowski1995time}.
In contrast, during inference, many modern ML models, such as convolutional neural networks (CNN), support vector machines (SVM) and logistic regression (LG) classifiers, rely on latent signal \textit{magnitude} to arrive at decisions \cite{maass1997networks}. We refer to these signal magnitudes, in the final layer of ML models, and before they are summed into scalars or logits for prediction, as \textit{classification embeddings (CE)}. The impact of signal magnitude on model decisions and generalization is often \textit{hidden} because the activation of individual latent features during inference is masked by summation and thresholding operations. 

An important problem of ML models is their perceived difficulty in generalizing to imbalanced data (see Supplemental Section S1 for a quantitative analysis) \cite{chawla2002smote,krawczyk2016learning,cao2019learning}. 
To that end, we investigate the importance of latent feature magnitude during the inference phase of supervised classifiers as a potential culprit in generalization for imbalanced data. We select three representative algorithms from the class of parametric ML models: CNN, SVM and LG. We demonstrate the importance of feature magnitude in parametric ML prediction and that these models rely on only a handful of features to predict classes for individual instances in both image and tabular data. 
In the case of imbalanced classes, the reliance on a few features with out-sized magnitudes prevents adequate generalization. This reliance on a few features for prediction persists even when classes with fewer instances are augmented through several widely used over-sampling techniques. 

In this paper, we make the following contributions to the study of imbalanced data:



\textbf{The importance of latent feature magnitude during inference.} 
We show that parametric ML models rely on a few, high magnitude latent features to predict individual class instances. This observation applies to both majority and minority classes. In the case of image data, a CNN requires approximately 12\% of classification embeddings to predict either majority or minority classes. 

\textbf{Minority class over-sampling does \textit{not} meaningfully change the ratio of latent features required for prediction.} Even though we augment minority classes with several over-sampling techniques, the number of latent features required for majority and minority class prediction do not meaningfully change. 


\textbf{Latent feature \textit{magnitude} is directly related to the \textit{frequency} with which the features appear in the training set for image data.} This relationship applies in the case of majority and minority classes, and with and without over-sampling. 

\textbf{The number of latent features required to predict an entire class is much larger than the number of features required to predict a single class instance.} This implies that parametric ML models learn different \textit{combinations} of latent features for a class. 

\section{Related Work}

Imbalanced learning focuses on how a disparity in the number of class instances affects the training of supervised classifiers. The classes are colloquially referred to as the majority class(es) (with more samples) and the minority class(es) (with fewer samples). Class numerical sample differences may be due to: (1) step imbalance (cliff effect disparity in class samples), or (2) exponential imbalance (a graduated difference in the number of instances in a multi-class setting).

Much of the research on generalization in parametric ML models has been conducted outside of imbalanced learning and has mainly focused on CNNs \cite{zhang2021understanding, arpit2017closer, neyshabur2014search, geirhos2020shortcut,shah2020pitfalls,neyshabur2017exploring}, where classifiers have difficulty generalizing to slightly harder images than those found in the original dataset \cite{recht2019imagenet} and to small changes in model inputs (adversarial examples), without explicit training \cite{hendrycks2019benchmarking, huber2021four,yalniz2019billion,kolesnikov2020big}. In contrast to these approaches, which mainly focus on weight regularization and data augmentation to improve generalization, we focus on the \textit{magnitude} of the feature embeddings learned by parametric ML models as the central culprit in generalization.

Within imbalanced learning, the reason why ML classifiers have difficulty generalizing to classes with fewer instances has been attributed to: numerical class imbalance \cite{krawczyk2016learning}, class overlap \cite{batista2004study,denil2010overlap,prati2004class,garcia2007empirical}, subconcepts, and disjuncts \cite{jo2004class,weiss2004mining}. Instead of searching for the problem of generalization in class overlap or numerical class imbalance, we focus on the \textit{summation} function in supervised ML models.

When training ML models with imbalanced data, common techniques that improve generalization with respect to the minority class include: cost-sensitive approaches that increase the penalty of minority class misclassification, ensemble methods, and oversampling the minority class, which is a form of data augmentation \cite{japkowicz2002class,he2009learning,krawczyk2016learning}. Oversampling balances the number of class training instances by synthetically increasing the number of minority class instances. 

Three representative oversampling techniques that use data augmentation (DA) are: SMOTE \cite{chawla2002smote}, ADASYN \cite{he2008adasyn}, and REMIX \cite{bellinger2020remix}. SMOTE and ADASYN draw samples from the same class (minority class) for augmentation; although ADASYN specifically samples hard-to-classify instances. REMIX integrates mixup \cite{zhang2017mixup}, which is widely used in balanced image training, into imbalanced learning. It samples instances for augmentation from both the majority and minority classes and uses label smoothing. SMOTE, ADASYN and REMIX all implement DA for class imbalance at the front-end of model training (i.e., on raw input features). 

In addition to these methods, we examine two DA techniques that improve generalization for minority classes that work on \textit{latent} features: EOS \cite{dablain2022efficient} and DSM \cite{dablain2023towards}. EOS draws samples from adversary classes to synthetically augment minority classes. DSM is based on DeepSMOTE \cite{dablain2022deepsmote}, except that it draws same class latent features from the model itself, instead of using a separate auto-encoder. 

\vspace{-.25cm}
\section{Classification Embedding Motivation}
\label{sec:exp}
\vspace{-.25cm}

To motivate the importance of signal magnitude to the classification process of parametric ML models, we first define the latent features, or classification embeddings (\textit{CE}), used in our experiments for LG, CNN, and SVM models. We describe how parametric ML models require that the magnitude of \textit{CE} for the predicted class be greater than the competing class(es). In this environment, it is possible for only a few latent features (\textit{CE}) to dominate class prediction, and hence limit model generalization capacity. To simplify our discussion, we assume that LG and SVM models perform binary classification and that CNNs perform multi-class classification. We also assume that inference is performed on each instance, instead of in batches.

\smallskip
\textbf{LG classifier.} During inference, a LG classifier can be described as:
\begin{equation} \label{eq:1}
 y = \phi (\sum_{i=0}^{F_D} CE_i + b)
\end{equation}
where y is a binary label ($y \in \{0,1 \}$), $\phi$ is a thresholding function described below, $F_D$ is the dimension of the features, $\sum CE_i$ is the summation of a vector of classification embeddings and \textit{b} is a bias term. 
 \textit{CE} can be described as:
\vspace{-.1cm}
\begin{equation}\label{eqn_ce}
 CE_i = FE_i \cdot W_i
\end{equation}
where the \textit{FE} is a vector of features  and \textit{W} is a vector of learned weights. 
For LG models, the real determinant of label prediction during inference is the \textit{summation} of the \textit{CE}. If the sum of \textit{ CE} plus bias is negative, then the sigmoid and rounding functions will produce a 0 value (class prediction). It will produce a value of 1 (class prediction) if the sum of the \textit{CE} plus bias term is positive. Thus, the prediction of the LG model is highly dependent on the linear sum of the magnitudes of features contained in a vector of \textit{CE}, plus a bias term, during inference.

\smallskip
\textbf{CNN classifier.}
The final classification layer in a CNN can be described by Equation~\ref{eq:1}; however, there are important differences in $\phi$ and how \textit{FE} are encoded for CNNs. For a CNN classification layer, the \textit{argmax} function can be substituted for $\phi$ so that Equation~\ref{eq:1} becomes:
\vspace{-.3cm}
\begin{equation}
 y = argmax(_{c=0}^{C}( \sum_{i=0}^{F_D} CE_i + b_i))_c
\end{equation}
where \textit{C} is the number of classes and the \textit{argmax} returns the index (label) of the class with the largest sum. In a CNN, the \textit{FE} are described as:
\vspace{-.1cm}
\begin{equation}
 FE = pool(f(\cdot))
\end{equation}
where \textit{pool} is a pooling function and $f(\cdot)$ is the embedding output of multiple convolution layers \cite{dablain2022understanding}. Although \textit{ FE} and related weights are determined through different operations in the LG and CNN models, the \textit{final prediction} is based on the \textit{summation of CE plus a bias term}.

\smallskip
\textbf{SVM classifier.} 
For SVMs, Equation~\ref{eq:1} can be rewritten as:
\vspace{-.2cm}
\begin{equation} 
 y = Sign (\sum_{i=0}^{F_D} CE_i + b)
\end{equation}
The Sign function returns a label based on whether the summation is negative (label: 0) or positive (label: 1). During the SVM inference process, the components of \textit{CE} - \textit{FE} and \textit{W} - are also computed differently. In a SVM classifier, \textit{FE} can be expressed as:
\vspace{-.05cm}
\begin{equation} 
 FE = K(SV,I_t)
\end{equation}
where \textit{SV} are support vectors, $I_t$ is a test instance, and \textit{K} is a kernel function. Kernel functions can consist of the dot product, polynomial functions, or radial basis functions, which are standard approaches to compute the similarity of two vectors.

\smallskip
\textbf{Conclusion.}
In all three models, prediction, and hence generalization, depends on the sum of latent features (\textit{CE}). In this environment, it is possible for only a handful of \textit{CE} to dominate, which may limit the ability of the models to generalize to instances not observed during training. This limitation is especially pronounced in the case of imbalanced learning because minority classes, which have a decreased number of training examples, do not contain a sufficient number of instances to ensure class feature diversity. In the following sections, we show that, even with data augmentation to supplement the number of minority class instances, parametric ML models continue to rely on a limited number of \textit{CE} for prediction, which can limit their generalization capacity when learning with imbalanced data.

\section{Experimental Study} 
We investigate the role of the magnitude of the latent feature (\textit{CE}) magnitude during the inference process of LG, SVM, and CNN models when learning with imbalanced data. We demonstrate, through our experiments, that regardless of \textit{where} DA is performed (e.g., at the front end of training or in latent space) to improve minority class generalization, ML models still rely on a limited number of \textit{CE} for prediction when learning with imbalanced data. We base our investigation on the following research questions (RQs):
\begin{itemize}
\setlength{\itemsep}{-1pt}

    \item \textbf{RQ1:} Do parametric ML models rely on a \textit{limited number} of latent features (\textit{CE}) to predict a \textit{single} instance? If models only rely on a handful of features for single instance predictions, do these features have larger \textit{magnitudes} than nonrelevant features?
    
    \item \textbf{RQ2:} Do parametric ML models rely on a larger number of \textit{CE} than in RQ1 to predict a \textit{ entire} class? Does this imply that these models rely on any linear combination of \textit{CE }that sum to a requisite threshold vs. learning invariant features that describe all class instances?
    
    \item \textbf{RQ3:} Is there a relationship between the latent characteristic \textit{ magnitude} and the \textit{frequency} that \textit{CE} occur in a training set? 
\end{itemize}

For a CNN, we use Resnet-32 and Resnet-56 architectures \cite{he2016deep}. We test the LG and SVM classifiers using tabular data and a CNN using image data. 
For tabular data, we select 5 imbalanced binary classification datasets: Ozone, Scene, Coil, Thyroid and US Crime from the UCI Machine Learning library \cite{Dua:2019}. 
For image data, we use 3 imbalanced datasets: CIFAR-10 \cite{krizhevsky2009learning}, Places \cite{zhou2017places}, and INaturalist \cite{van2018inaturalist}. 
For tabular data, we compare a baseline model trained with imbalanced data with 3 methods that employ DA at the front end of model training (SMOTE, ADASYN, and REMIX). For image data, we compare a baseline model trained with imbalanced data with a method (REMIX) that employs DA at the front-end of model training and two that implement DA in latent space (DSM and EOS).


\begin{table}[h!]
\vspace{-.7cm}
\footnotesize
\caption{Image Datasets}
\label{tab:ch8_img_data}
\begin{tabular}{ p{1.8cm}p{1.6cm}p{1.6cm}p{1.3cm}
p{1.2cm}p{1cm}p{1cm}p{1cm}}
\toprule

\textbf{Dataset} & \textbf{\# Class} &
\textbf{Imbal Type} &
\textbf{Max \mbox{Imbal} Level} & \textbf{\# Train Maj} &
\textbf{\# Train Min} &
\textbf{\# \mbox{Test} \mbox{Maj}} &
\textbf{\mbox{\#} Test \mbox{Min}}\\

\midrule

CIFAR-10 & 10 & Expon & 100:1 & 5000 & 50 & 1000 & 1000\\

Places & 5 & Step & 5:1 & 2500 & 500 & 250 & 250\\

INaturalist & 5 & Step & 5:1 & 6250 & 1250 & 500 & 250\\

\bottomrule

\end{tabular}
\vspace{-.5cm}
\end{table}

\subsection{Image data details}

This section provides details of the three image datasets: CIFAR-10 \cite{krizhevsky2009learning}, Places \cite{zhou2017places}, and INaturalist \cite{van2018inaturalist}. The number of class instances in CIFAR-10 is initially balanced, and we imbalance them exponentially, with a 100:1 maximum level. We follow Cao et al.'s \cite{cao2019learning} exponential imbalance formula for CIFAR-10 (5000, 2997, 1796, 1077, 645, 387, 232, 139, 83, 50). We select five classes from the Places dataset: airfield, amusement park, acquarium, baseball field and barn. The number of class instances is also initially balanced. We employ 10:1 step imbalance (airfield and amusement park with 5K training samples and the rest with 500). We randomly select samples from five classes with the INaturalist dataset: plant, insect, bird, reptile and mammal, with 5:1 step imbalance (plant and insect are the majority classes; bird, reptile and mammal are the minority). We use datasets with 5 to 10 classes because fewer classes allows us to visualize feature distributions. We report our results based on the average of three random permutations of the training sets. Table~\ref{tab:ch8_img_data} contains more details of the image training and test sets.


\subsection{Tabular data details}

For tabular data, our dataset selection criteria are: imbalance ratio greater than 10:1, number of samples greater than 1,000, number of features greater than 50, and datasets containing only non-categorical features (i.e., integer and real numbers). The key details of the datasets are summarized in Table~\ref{tab: ch8_data}. We report our results based on the average of 5 randomly drawn splits of the datasets, with a 70:30 training to test split. We select our datasets from the UCI Machine Learning Library \cite{Dua:2019}

\begin{table}[h!]
\vspace{-.7cm}
\footnotesize
\caption{\textbf{Tabular Datasets}}
\label{tab: ch8_data}
\begin{tabular}{ p{1.9cm}p{1.9cm}p{2.1cm}p{2.1cm}}
\toprule

\textbf{Dataset} & \textbf{Imbalance Ratio} &
\textbf{\# \mbox{Samples}} &
\textbf{\# \mbox{Features}} \\

\midrule

Ozone & 34:1 & 2,536 & 72 \\
Scene & 13:1 & 2,407 & 294 \\
Coil & 16:1 & 9,822 & 85 \\
Throid & 15:1 & 3,772 & 52 \\
US-crime & 12:1 & 1,994 & 100 \\
\bottomrule

\end{tabular}
\end{table}
\vspace{-.7cm}

\begin{figure*}[t!]
   \vspace{-.8cm}
  \centering
   \subfloat[CIFAR-10]{\includegraphics[width=0.33\textwidth]{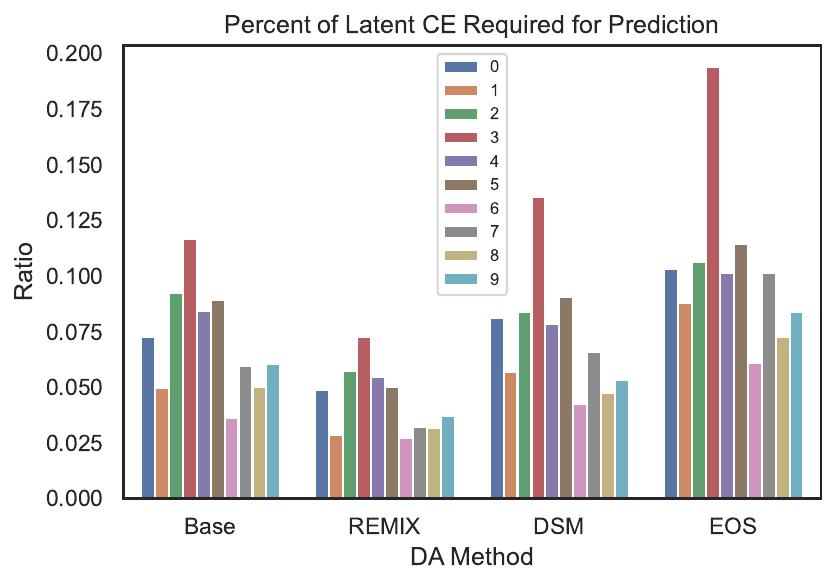}}
   \hfill
\subfloat[INaturalist]{\includegraphics[width=0.33\textwidth]{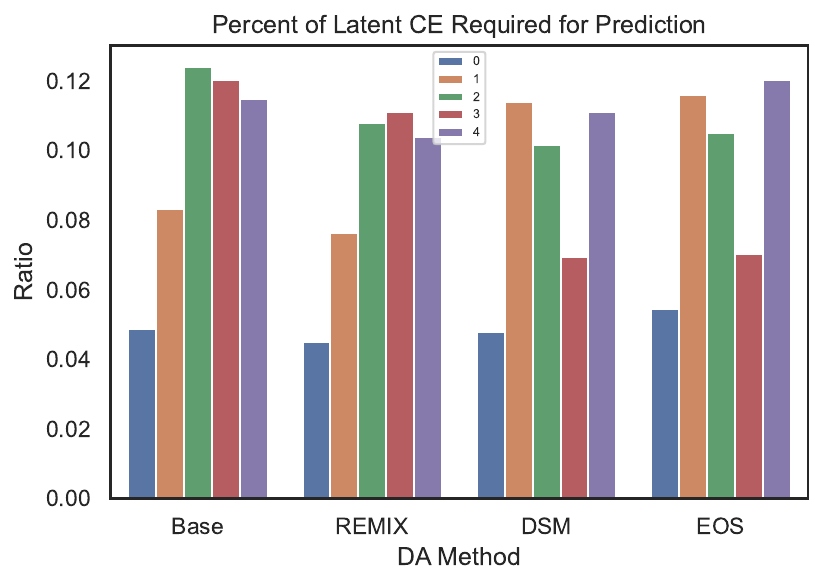}}
  \hfill
  \subfloat[Places]{\includegraphics[width=0.33\textwidth]{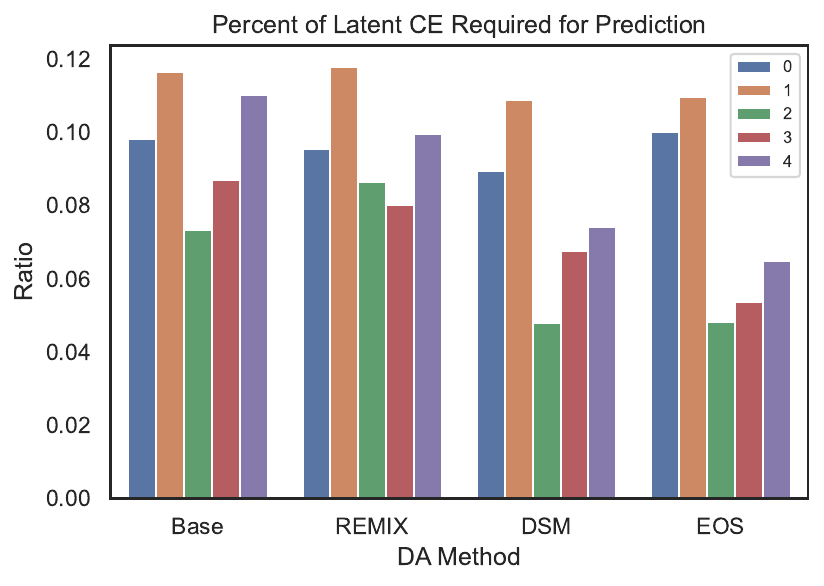}}
  \caption{This figure displays the percentage of \textit{CE} (y-axis) that are required to predict a label for a single instance for 3 image datasets. Classes with more samples (majority) have lower class labels (e.g., 0), with the imbalance increasing from left to right, depending on exponential or step. A base model is trained with imbalanced data and 3 models are trained with over-sampling methods (x-axis). With a few exceptions, 12\% or fewer \textit{CE} are needed to predict a single instance of a minority or majority class. This holds whether the models are trained with or without minority class DA.}
  \label{fig_tp_count_img}
  \vspace{-.3cm}
\end{figure*}

\subsection{Model training}
For LG and SVM (RBF kernel) models, we use standard classifiers that are publicly available from the SK Learn library with default hyperparameters \cite{scikit-learn}. For CNN training, we employ the CNN training regime for imbalanced data established by Cao et al. \cite{cao2019learning}, with the following modifications. For CIFAR-10, we train for 200 epochs and for Places and INaturalist, we train for 40 epochs. We use a standard Resnet-32 architecture \cite{he2016deep} for CIFAR-10 and a standard Resnet-56 architecture for Places and INaturalist. We use a single NVIDIA 3070 GPU with PyTorch \cite{paszke2017automatic}. 

\section{Results}

\subsection{\textbf{RQ1: Number of latent features to predict an instance} }

Figure~\ref{fig_tp_count_img} shows the number of \textit{CE} that are required to predict true positive (TP) instances in 3 image datasets. The results are shown as a percentage of the dimension of the classification embeddings (e.g., dimension of 64 for a Resnet-32). For this purpose, the number of CE needed to exceed the sum of CE of the next largest predicted class for each TP instance was determined. The average number of required \textit{CE} was then calculated based on 3 splits of each dataset.

\begin{figure}[t!]
   \vspace{-.5cm}
  \centering
\subfloat[Ozone]{\includegraphics[width=0.33\textwidth]{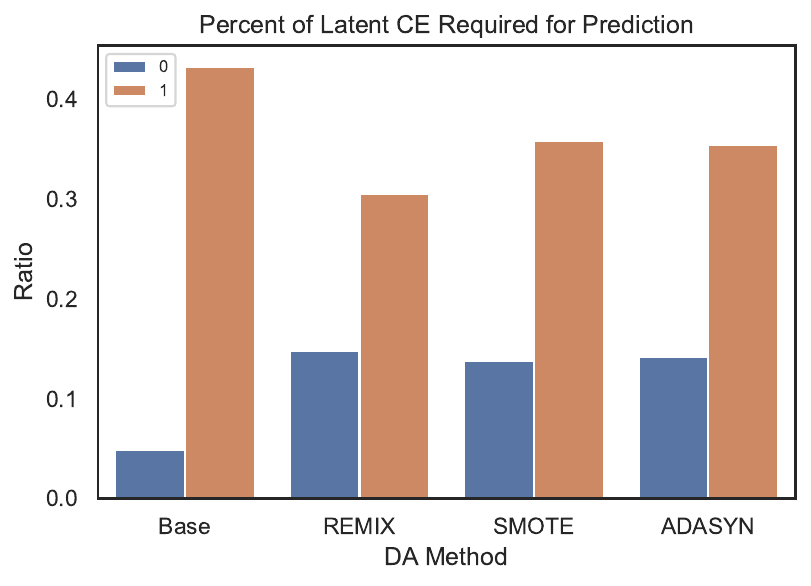}}
  \hfill
  \subfloat[Scene]{\includegraphics[width=0.33\textwidth]{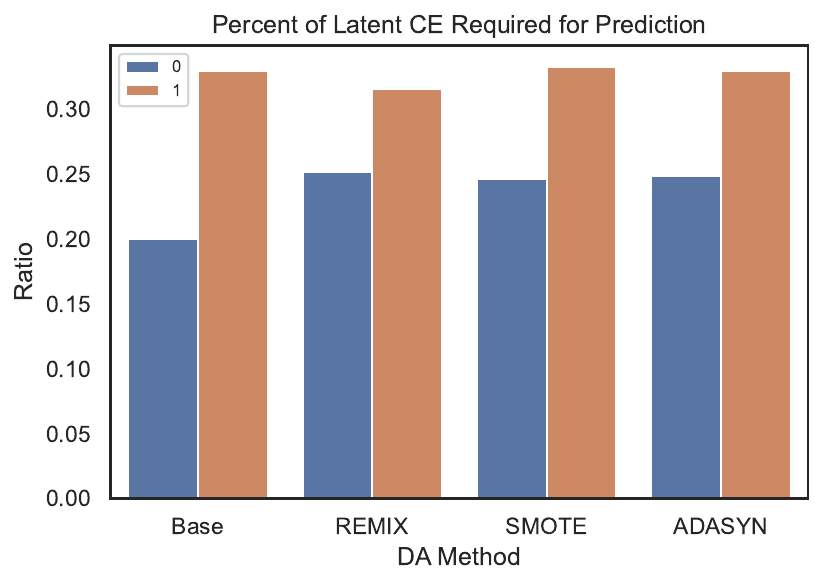}}
  \hfill
  \subfloat[Coil]{\includegraphics[width=0.33\textwidth]{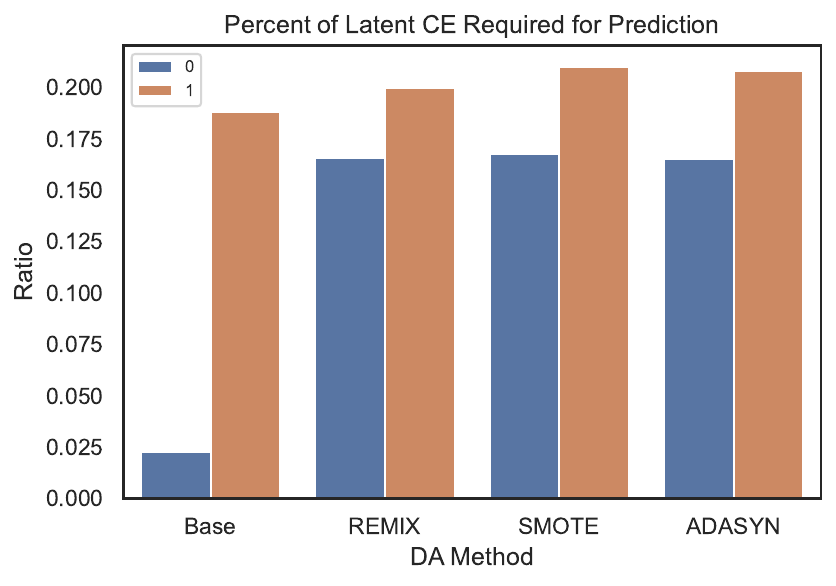}}
  \hfill
  \subfloat[Thyroid]{\includegraphics[width=0.49\textwidth]{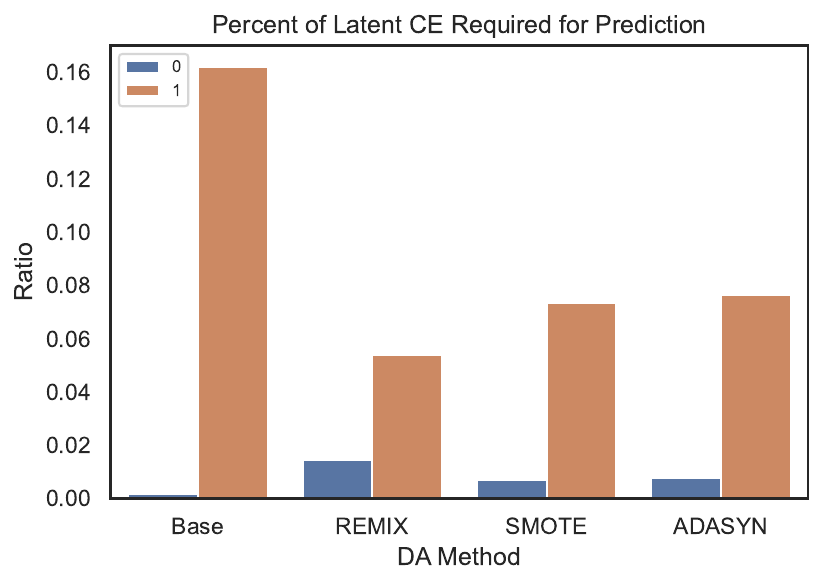}}
  \hfill
  \subfloat[US Crime]{\includegraphics[width=0.49\textwidth]{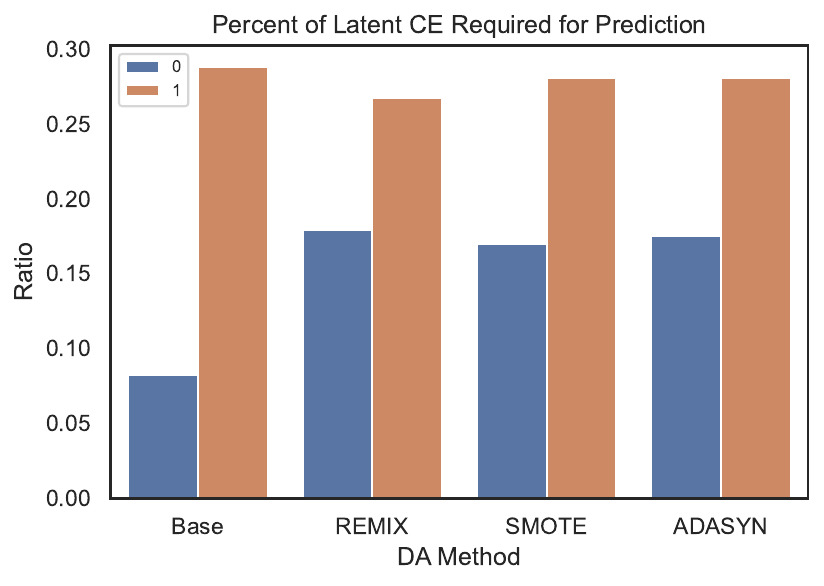}}
  \caption{This figure displays the percentage of \textit{CE} (y-axis) required to predict a \textit{single instance} for LG models on 5 tabular datasets.  Class 0 (blue) denotes the majority and class 1 (brown) is the minority. A base model is trained on imbalanced data and 3 models are trained with over-sampling. For all datasets, more \textit{CE} are required to predict minority classes, both with and without over-sampling.}
  \label{fig_tp_count_lg}
  \vspace{-.6cm}
\end{figure}

In Figure~\ref{fig_tp_count_img}, each class is represented by a different color bar. 
The bars (classes) are grouped by DA method, with a base model trained on imbalanced data and 3 models for DA methods (REMIX, DSM and EOS). 
In all cases, only a small percentage of \textit{CE} are required to predict a label for an instance of a class. With the exception of a single class in CIFAR-10 (class \#3 - cats), fewer than 12\% of \textit{CE} are required to predict a TP instance for a class, when averaged over three splits for each method and each dataset. These results hold whether models are trained on imbalanced or augmented data, and whether DA occurs in real or latent space. 

Figure~\ref{fig_tp_count_lg} shows the percentage of latent features (\textit{CE}) that are required to correctly predict an instance in 5 tabular datasets using LG models. Here, a model is trained on baseline, imbalanced data and with 3 DA methods (SMOTE, ADASYN and REMIX). Overall, the percentage of \textit{CE} required to predict an instance ranges from a low of 2.5\% to just over 40\%. We hypothesize that LG models generally require more \textit{CE} to predict an instance than CNN models because deep CNNs learn a more compact embedding of the input than single layer LG models. 

For all datasets in the case of LG classifiers, more \textit{CE} are required to correctly predict minority class instances (class \#1, with brown bars) than majority class instances (class \#0, with blue bars). 

In the case of base models trained on imbalanced data, there is a wide gap; however, DA methods narrow the gap such that the number of \textit{CE} required to predict the two classes is more evenly balanced. We hypothesize that the reason why over-sampling narrows this gap is because single layer LG models have more difficulty predicting majority classes when the number of minority class instances is augmented; hence, they require more \textit{CE} for majority class instances.

\begin{figure}[t!]
   \vspace{-.4cm}
  \centering
\subfloat[Ozone]{\includegraphics[width=0.33\textwidth]{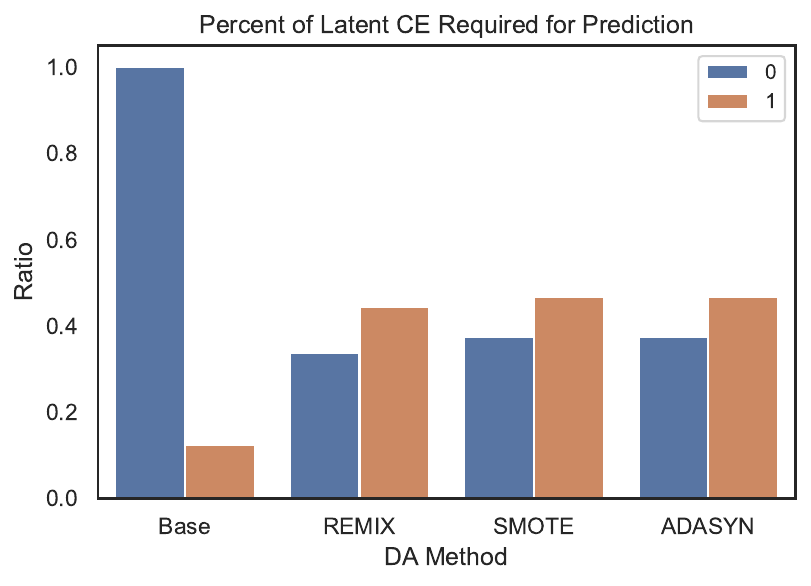}}
  \hfill
  \subfloat[Scene]{\includegraphics[width=0.33\textwidth]{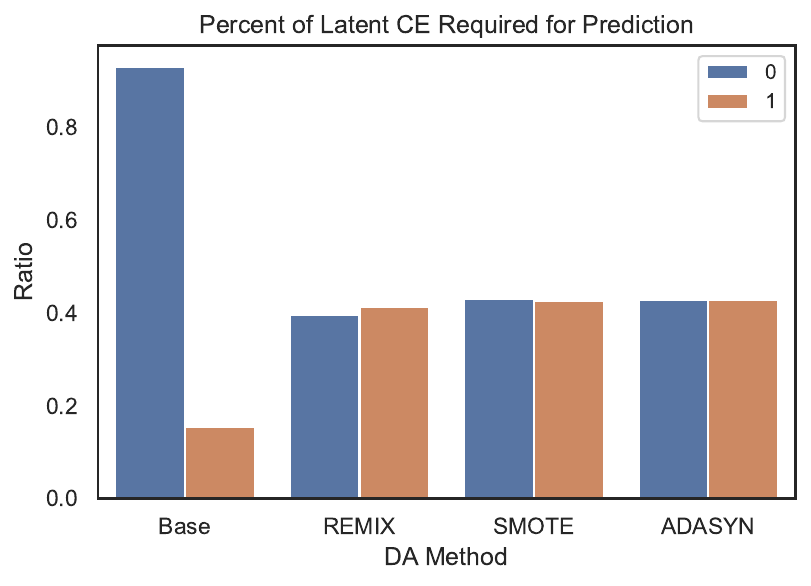}}
  \hfill
  \subfloat[Coil]{\includegraphics[width=0.33\textwidth]{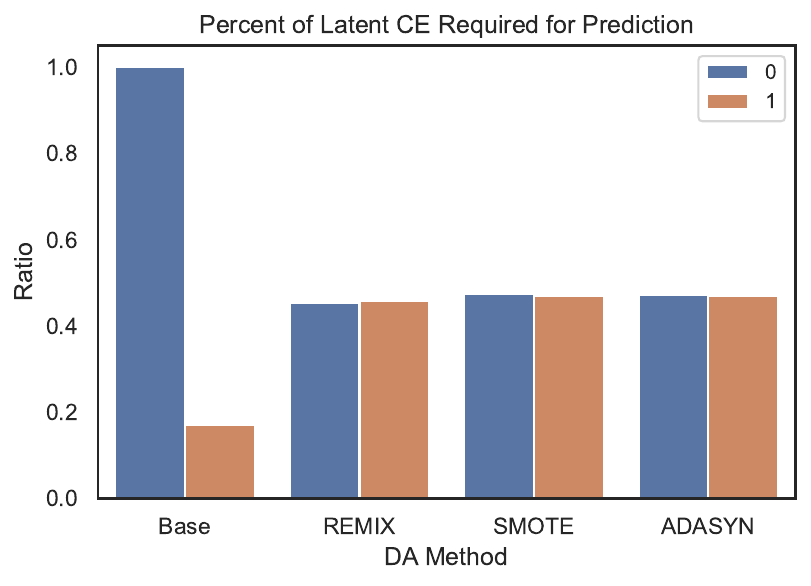}}
  \hfill
  \subfloat[Thyroid]{\includegraphics[width=0.49\textwidth]{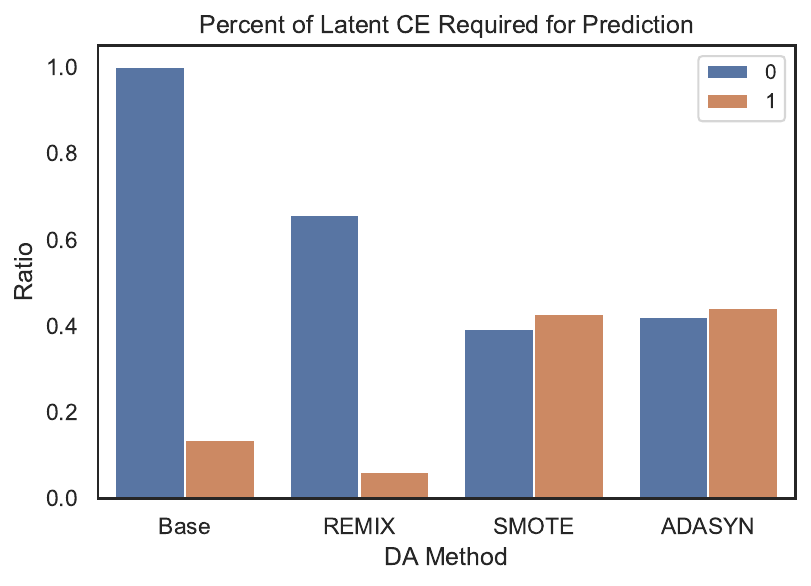}}
  \hfill
  \subfloat[US Crime]{\includegraphics[width=0.49\textwidth]{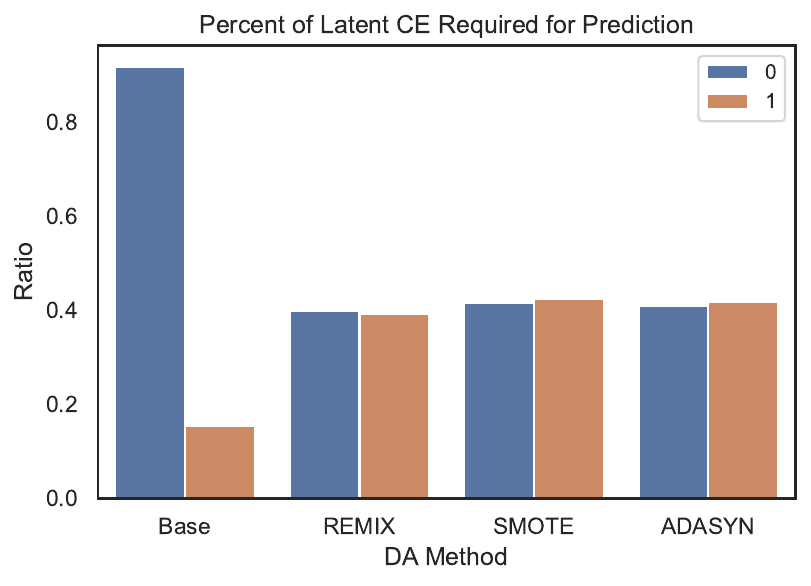}}
  \caption{This figure displays the percentage of \textit{CE} required to predict a single instance using SVM models for 5 UCI tabular datasets. Blue bars denote majority, and brown bars denote minority, classes. When training with base (imbalanced) data, majority class instances require, on average, almost all \textit{CE} for prediction; however, when trained with SMOTE and ADASYN, and in some cases with REMIX, the percentage of \textit{CE} falls to approx. 40\% for both majority and minority classes.}
  \label{fig_tp_count_sv}
  \vspace{-.5cm}
\end{figure}

Figure~\ref{fig_tp_count_sv} displays the percentage of \textit{CE} that are required, on average, to correctly predict a class using SVM models for 5 tabular datasets. The models are trained with base, imbalanced data and 3 different DA methods. In contrast to the CNN and LG examples presented above, SVM models rely on a much larger percentage of \textit{CE} (and support vectors) to correctly predict class labels. In the base, imbalanced models, nearly 100\% of \textit{CE} are required to predict the majority class, which may be indicative of a high degree of memorization and a less efficient encoding.

In SVM models, with the exception of REMIX in the Thyroid dataset, DA causes the number of \textit{CE} required to predict majority and minority classes to be more evenly balanced - at $\approx$40\% for models trained with DA.

\begin{figure*}[b!]
   \vspace{-.5cm}
  \centering
   \subfloat[CIFAR-10]{\includegraphics[width=0.33\textwidth]{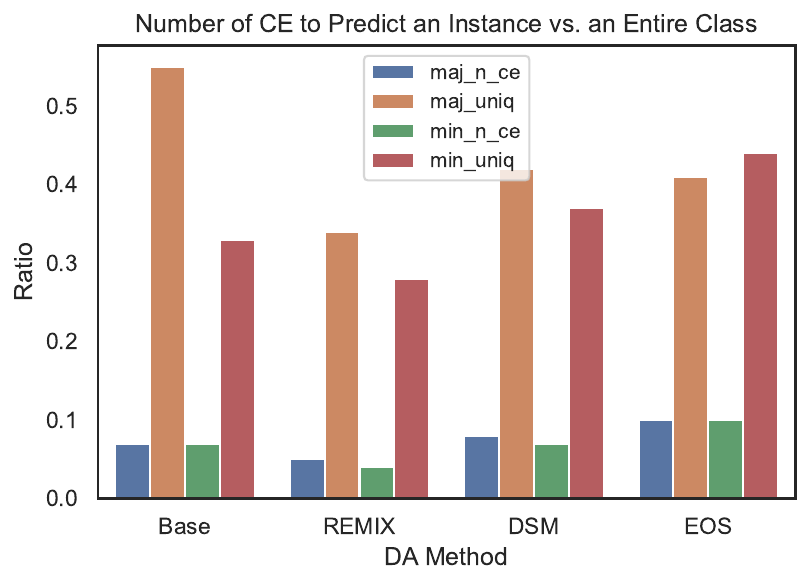}}
   \hfill
\subfloat[INaturalist]{\includegraphics[width=0.33\textwidth]{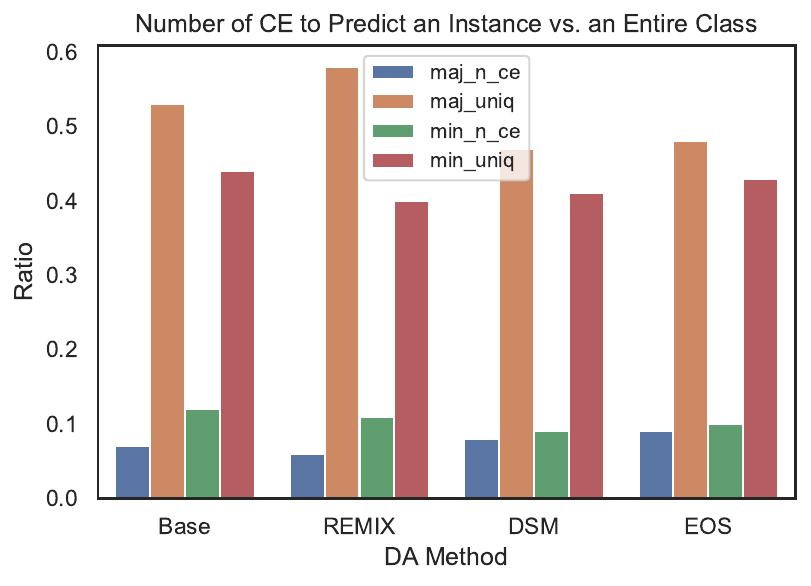}}
  \hfill
  \subfloat[Places]{\includegraphics[width=0.33\textwidth]{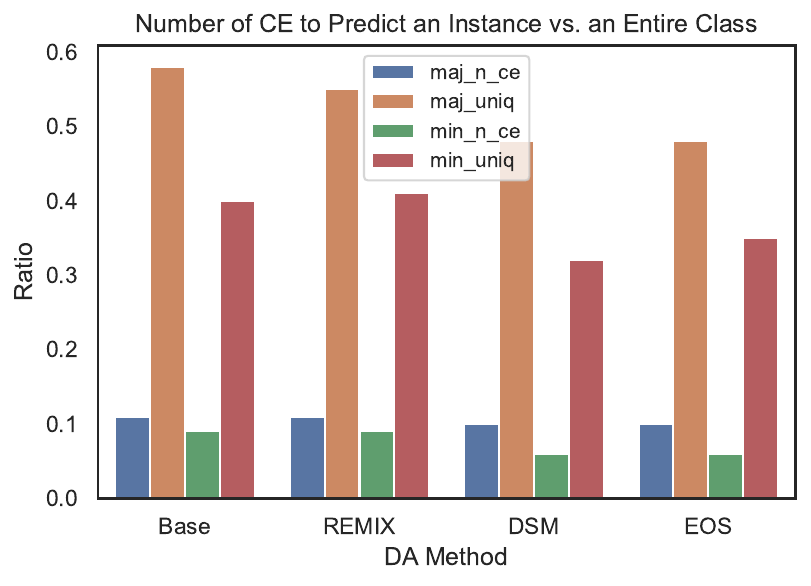}}
  \hfill

  \caption{This figure compares the number of \textit{CE} or features required to predict an \textit{instance} versus the number of \textit{CE} required to predict an \textit{entire class}. }
  \label{fig_ce_inst_cls}
  \vspace{-.3cm}
\end{figure*}

\begin{figure}[t!]
   \vspace{-.4cm}
  \centering
\includegraphics[width=0.99\textwidth]{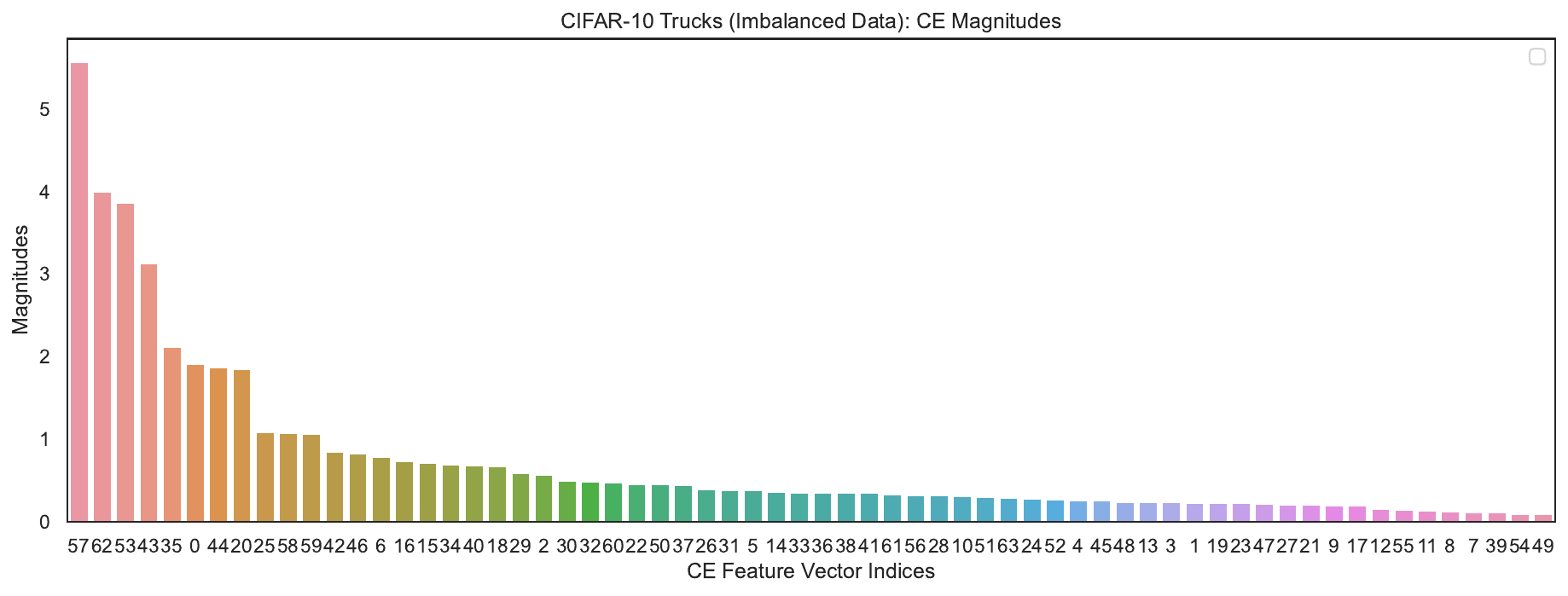}
  
  \caption{This figure illustrates that, out of a total of 64 latent features in a Resnet-32 model trained to predict the CIFAR-10 truck class, only a handful of features have large feature magnitudes (y-axis). This figure depicts the minority class (class \#9 - trucks). The indices of the latent features are shown on the x-axis.}
  \label{fig_trk_mags}
  \vspace{-.5cm}
\end{figure}

Figure~\ref{fig_ce_inst_cls} compares the number of \textit{CE} required to predict a single instance versus the number of \textit{CE} required to predict an entire class. The number of \textit{CE} to predict an entire class is the number of unique \textit{CE} indices compiled from the top-10 \textit{CE} for each TP instance in a class. 
For multi-class datasets, such as CIFAR-10, Places and INaturalist, we split the datasets into the majority class(es) and minority class(es). The majority class is the class with the most instances (two classes in the case of Places and INaturalist, which have step imbalance and one class in the case of CIFAR-10). The minority class(es) are all other classes. In cases where multiple classes are combined, we average the number of \textit{CE} required to make predictions. We express the number of \textit{CE} as a percentage of the total dimension of the classification embeddings.
In the figure, the number of \textit{CE} required to predict a majority instance is denoted in blue and a minority instance in green. The number of unique latent features (\textit{CE}) needed to predict \textit{all} instances in a class is denoted in orange for majority classes and red for minority classes. In all cases, a relatively low number of \textit{CE} are required to predict an individual class \textit{instance} (blue and green for majority and minority classes, respectively) and more are required to define \textit{all} class instances (orange and red for majority and minority classes, respectively). This implies that the \textit{identity} of the features (\textit{CE}) required to predict individual instances varies across all class instances. 

\subsection{\textbf{RQ2: Importance of feature magnitude} }

\begin{figure*}[t!]
  \centering
\subfloat[Image - CNN]{\includegraphics[width=0.49\textwidth]{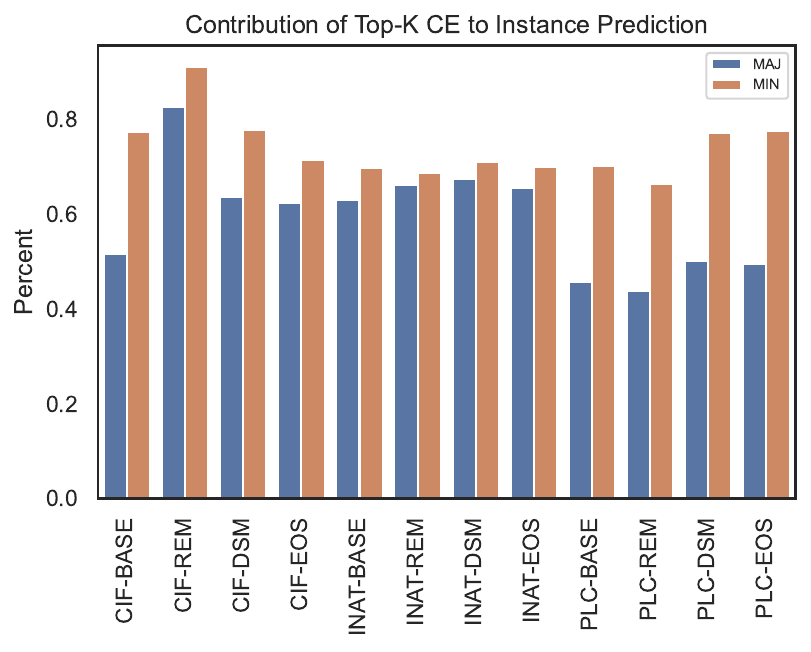}}
  \hfill
  \subfloat[Tabular - LG]{\includegraphics[width=0.49\textwidth]{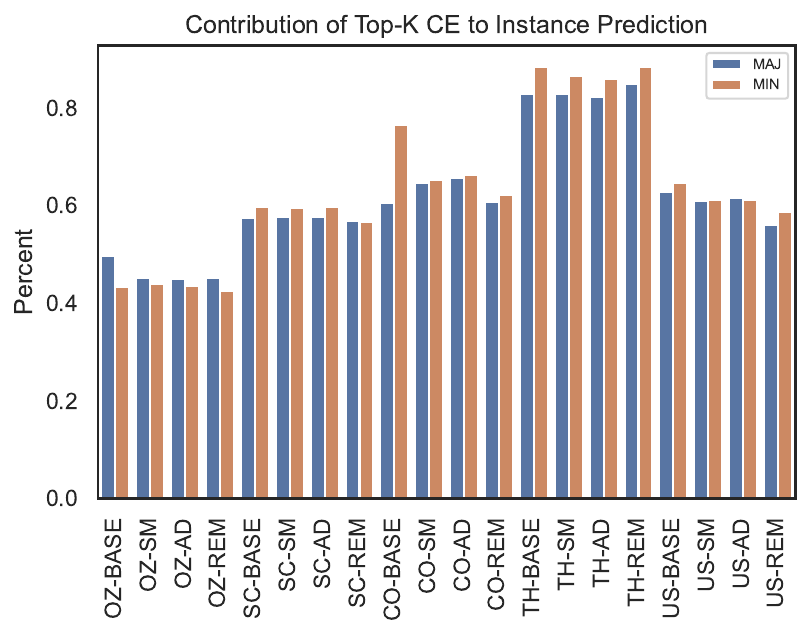}}
  \hfill
  \subfloat[Tabular - SVM]{\includegraphics[width=0.79\textwidth]{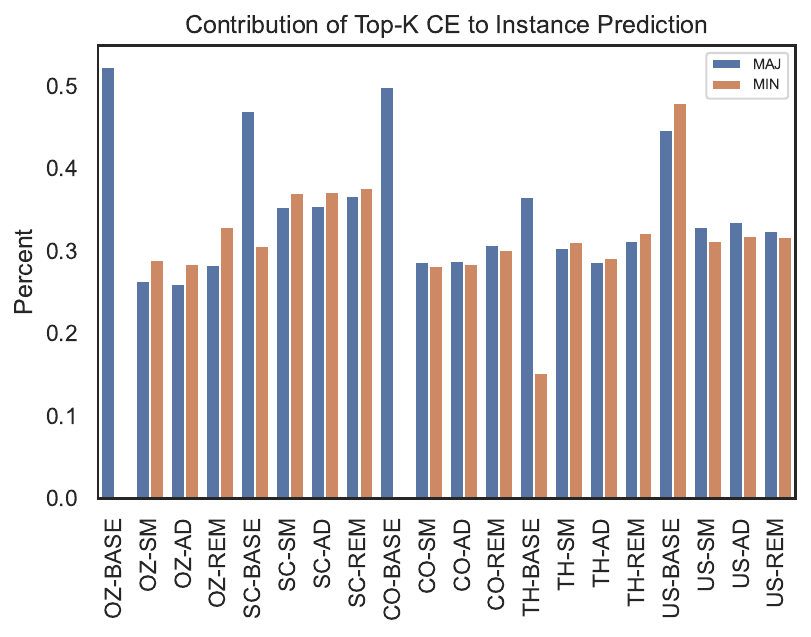}}
  
  \caption{This figure displays the contribution of the top-10\% of \textit{CE} with the largest average magnitudes. Here, the majority class(es) are blue and minority class(es) are orange. On the x-axis, the models trained with the respective data are presented and the ratio of the contribution of the top-10\% of \textit{CE} with the largest average magnitudes to the total average class \textit{CE} is shown on the y-axis.}
  \label{fig_topk_mags}
  \vspace{-.7cm}
\end{figure*}

Figure~\ref{fig_trk_mags} visualizes the importance of feature magnitude to label prediction during the inference process of parametric ML models. 
Here, the mean magnitudes averaged across all class instances represent the output of the feature maps that serve as latent input to the final classification layer in a CNN. In this figure, the model is trained with imbalanced CIFAR-10 data (trucks are the class with the fewest number of training instances). The mean magnitudes of the pooled feature maps are sorted to show their relative size and are averaged across all instances in the truck class. We can see that the magnitude of only a few feature indices dominate, with the vector locations of the features listed on the x-axis.

\begin{figure}[t!]
  \centering
\subfloat[Imbal.: Frequency]{\includegraphics[width=0.49\textwidth]{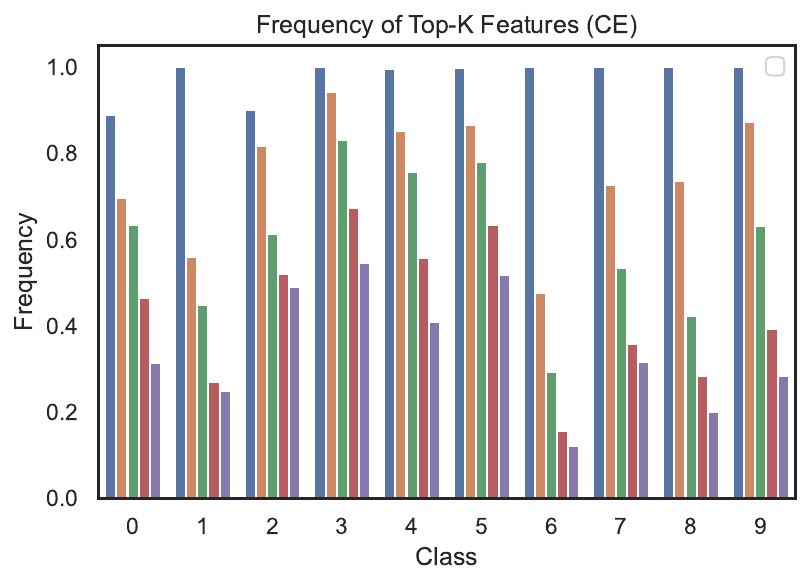}}
  \hfill
  \subfloat[Imbal.: Magnitude]{\includegraphics[width=0.49\textwidth]{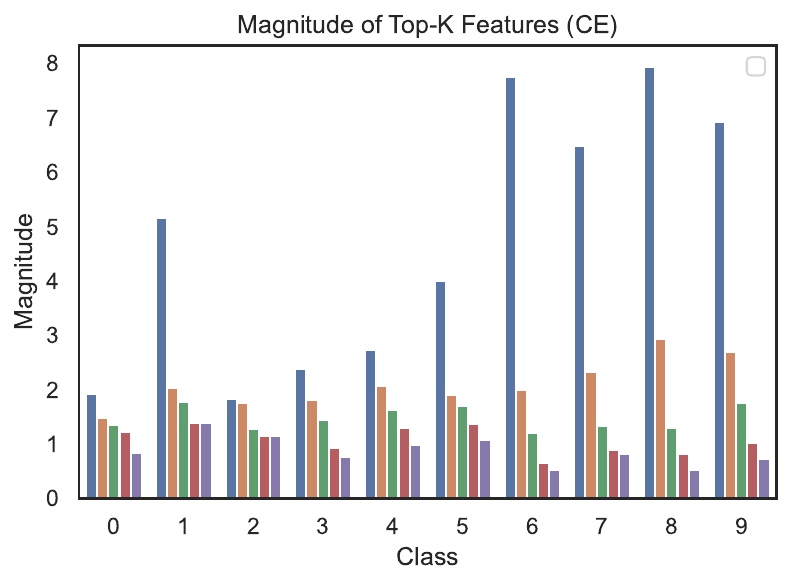}}
  \hfill
   \subfloat[EOS: Frequency]{\includegraphics[width=0.49\textwidth]{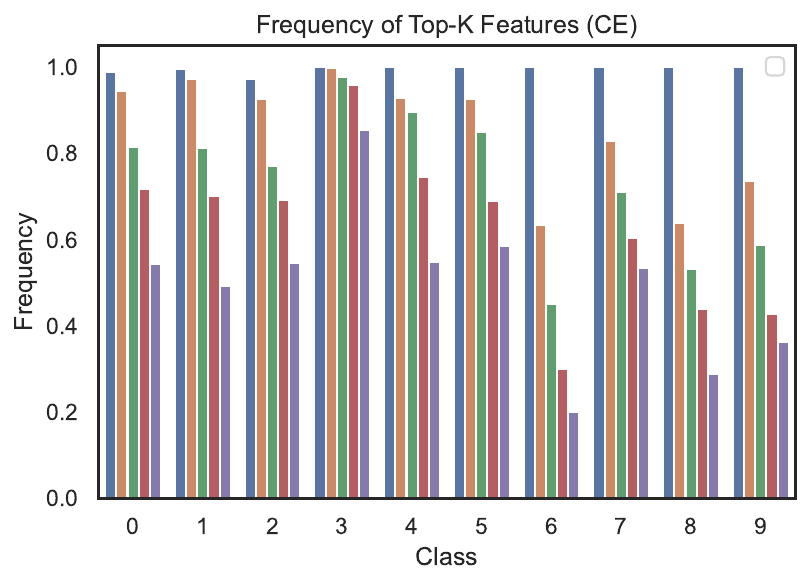}}
   \hfill
  \subfloat[EOS: Magnitude]{\includegraphics[width=0.49\textwidth]{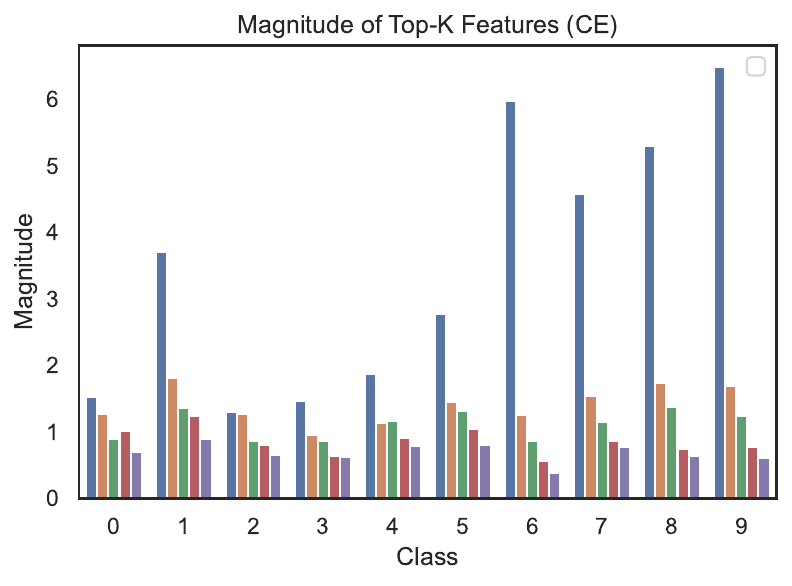}}
 
  \caption{This figure illustrates the relationship between the frequency with which features (\textit{CE}) are used to predict instances in a class and their magnitude for CIFAR-10. It illustrates a base model trained with imbalanced data and a model trained with augmented minority latent features using EOS. It shows the top-5 features with the highest mean magnitude for each class and the corresponding frequency with which those \textit{CE} appear in the training set. For all classes and with and without over-sampling, there is a clear relationship between magnitude and frequency - the magnitude of the feature declines as the frequency of that feature declines.} 
  \label{fig_n_app_mago}
  \vspace{-.4cm}
\end{figure}

Figure~\ref{fig_topk_mags} shows the percentage that the \textit{CE} with the largest magnitudes contribute to correct predictions (true positives). For CNN classification, the percentage is expressed as the sum of the top-10\% of \textit{CE} (i.e., the 10\% of \textit{CE} with the largest magnitude) divided by the sum of all positively signed \textit{CE} (i.e., \textit{CE} with a magnitude greater than zero). For LG and SVM models performing binary classification, the percentage is expressed as the sum of the top-10\% of \textit{CE} divided by the sum of the appropriately signed \textit{CE} (i.e., negative for 0 label and positive for 1 label). We report the percentages for models trained with imbalanced data (base) and DA methods.

For the LG and CNN models, on average, the top-10\% of \textit{CE} contributed between 40\% and over 80\% toward the class prediction (summation). For all image datasets, the average contribution of the top-10\% of \textit{CE} was 66.7\% and for LG models trained on tabular data, the average was 62.8\%. These percentages include both models trained with imbalanced data and synthetically augmented data. In the case of SVM models, the percentages were lower, with an average of 31.6\% for the top-10 \textit{CE}.

These results confirm that the final prediction in these models is based on the propagation of high-valued (magnitude) signals through one-layer (LG and SVM) or multi-layer (CNN) networks. Only a handful of features matter to the final prediction and those features have larger magnitudes than the features that contribute little to the final prediction. 


\subsection{\textbf{RQ3: Feature frequency \& magnitude}}

\

Figure~\ref{fig_n_app_mago} illustrates the relationship between the frequency with which \textit{CE} are used to predict instances in a class and their magnitude. For CIFAR-10, we show the frequency and the magnitude of the top-5 \textit{CE} for each class for a model trained with imbalanced data. The \textit{CE} with the highest frequency are sorted and displayed and then the magnitudes of the respective \textit{CE} are shown.
In the case of CIFAR-10 \textit{CE}, there is a clear correspondence between frequency and magnitude. In other words, the \textit{CE}, or encoded features, that occur with the highest frequency in each class also have the largest magnitudes. There is a clear matching of declining frequency and declining magnitude. In the figure, this relationship exists in CNNs trained with imbalanced and augmented data (EOS).

 This relationship is evident in other tested image datasets (Places and INaturalist) and DA methods; but not in tabular data. 
For more details, see 
\url{https://github.com/dd1github/Hidden_Influence_Magnitude}.

\vspace{-.2cm}
\section{Conclusion}

In this paper, we investigate the importance of latent signal magnitude to generalization. We show that generalization depends on the linear summation of latent features in classification layers. In this environment, it is possible for only a few latent features to dominant class prediction, which can limit the ability of ML models to generalize for classes with fewer, less diverse, training examples. Thus, the difficulty that parametric ML models have with minority class generalization can be traced, in part, to the inherent nature in which these models predict. 
 Through our experiments, this study contributes an important, although often overlooked and hidden, factor that limits the ability of parametric ML models to generalize in the face of class imbalance. 

\printbibliography

@article{zhang2017mixup,
  title={mixup: Beyond empirical risk minimization},
  author={Zhang, Hongyi and Cisse, Moustapha and Dauphin, Yann N and Lopez-Paz, David},
  journal={arXiv preprint arXiv:1710.09412},
  year={2017}
}

@article{chawla2002smote,
  title={SMOTE: synthetic minority over-sampling technique},
  author={Chawla, Nitesh V and Bowyer, Kevin W and Hall, Lawrence O and Kegelmeyer, W Philip},
  journal={Journal of artificial intelligence research},
  volume={16},
  pages={321--357},
  year={2002}
}

@inproceedings{he2008adasyn,
  title={ADASYN: Adaptive synthetic sampling approach for imbalanced learning},
  author={He, Haibo and Bai, Yang and Garcia, Edwardo A and Li, Shutao},
  booktitle={2008 IEEE international joint conference on neural networks (IEEE world congress on computational intelligence)},
  pages={1322--1328},
  year={2008},
  organization={IEEE}
}

@article{krizhevsky2009learning,
  title={Learning multiple layers of features from tiny images},
  author={Krizhevsky, Alex and Hinton, Geoffrey and others},
  year={2009},
  publisher={Citeseer}
}

@article{zhou2017places,
  title={Places: A 10 million Image Database for Scene Recognition},
  author={Zhou, Bolei and Lapedriza, Agata and Khosla, Aditya and Oliva, Aude and Torralba, Antonio},
  journal={IEEE Transactions on Pattern Analysis and Machine Intelligence},
  year={2017},
  publisher={IEEE}
}

@inproceedings{van2018inaturalist,
  title={The inaturalist species classification and detection dataset},
  author={Van Horn, Grant and Mac Aodha, Oisin and Song, Yang and Cui, Yin and Sun, Chen and Shepard, Alex and Adam, Hartwig and Perona, Pietro and Belongie, Serge},
  booktitle={Proceedings of the IEEE conference on computer vision and pattern recognition},
  pages={8769--8778},
  year={2018}
}

@inproceedings{he2016deep,
  title={Deep residual learning for image recognition},
  author={He, Kaiming and Zhang, Xiangyu and Ren, Shaoqing and Sun, Jian},
  booktitle={Proceedings of the IEEE conference on computer vision and pattern recognition},
  pages={770--778},
  year={2016}
}

@misc{Dua:2019 ,
author = "Dua, Dheeru and Graff, Casey",
year = "2017",
title = "{UCI} Machine Learning Repository",
url = "http://archive.ics.uci.edu/ml",
institution = "University of California, Irvine, School of Information and Computer Sciences" }

@article{abeles1993spatiotemporal,
  title={Spatiotemporal firing patterns in the frontal cortex of behaving monkeys},
  author={Abeles, M and Bergman, Hagai and Margalit, E and Vaadia, Eilon},
  journal={Journal of neurophysiology},
  volume={70},
  number={4},
  pages={1629--1638},
  year={1993},
  publisher={American Physiological Society Bethesda, MD}
}

@article{maass1997networks,
  title={Networks of spiking neurons: the third generation of neural network models},
  author={Maass, Wolfgang},
  journal={Neural networks},
  volume={10},
  number={9},
  pages={1659--1671},
  year={1997},
  publisher={Elsevier}
}

@article{sejnowski1995time,
  title={Time for a new neural code?},
  author={Sejnowski, Terrence J},
  journal={Nature},
  volume={376},
  pages={21--22},
  year={1995},
  publisher={Springer}
}

@article{dablain2022efficient,
  title={Efficient augmentation for imbalanced deep learning},
  author={Dablain, Damien and Bellinger, Colin and Krawczyk, Bartosz and Chawla, Nitesh},
  journal={IEEE 39th International Conference on Data Engineering},
  year={2023}
}

@article{zhang2021understanding,
  title={Understanding deep learning (still) requires rethinking generalization},
  author={Zhang, Chiyuan and Bengio, Samy and Hardt, Moritz and Recht, Benjamin and Vinyals, Oriol},
  journal={Communications of the ACM},
  volume={64},
  number={3},
  pages={107--115},
  year={2021},
  publisher={ACM New York, NY, USA}
}

@inproceedings{arpit2017closer,
  title={A closer look at memorization in deep networks},
  author={Arpit, Devansh and Jastrzebski, Stanislaw and Ballas, Nicolas and Krueger, David and Bengio, Emmanuel and Kanwal, Maxinder S and Maharaj, Tegan and Fischer, Asja and Courville, Aaron and Bengio, Yoshua and others},
  booktitle={International conference on machine learning},
  pages={233--242},
  year={2017},
  organization={PMLR}
}

@article{hendrycks2019benchmarking,
  title={Benchmarking neural network robustness to common corruptions and perturbations},
  author={Hendrycks, Dan and Dietterich, Thomas},
  journal={arXiv preprint arXiv:1903.12261},
  year={2019}
}

@inproceedings{kolesnikov2020big,
  title={Big transfer (bit): General visual representation learning},
  author={Kolesnikov, Alexander and Beyer, Lucas and Zhai, Xiaohua and Puigcerver, Joan and Yung, Jessica and Gelly, Sylvain and Houlsby, Neil},
  booktitle={Computer Vision--ECCV 2020: 16th European Conference, Glasgow, UK, August 23--28, 2020, Proceedings, Part V 16},
  pages={491--507},
  year={2020},
  organization={Springer}
}

@article{yalniz2019billion,
  title={Billion-scale semi-supervised learning for image classification},
  author={Yalniz, I Zeki and J{\'e}gou, Herv{\'e} and Chen, Kan and Paluri, Manohar and Mahajan, Dhruv},
  journal={arXiv preprint arXiv:1905.00546},
  year={2019}
}

@inproceedings{huber2021four,
  title={A four-year-old can outperform resnet-50: out-of-distribution robustness may not require large-scale experience},
  author={Huber, Lukas S and Geirhos, Robert and Wichmann, Felix A},
  booktitle={SVRHM 2021 Workshop@ NeurIPS},
  year={2021}
}

@article{geirhos2020shortcut,
  title={Shortcut learning in deep neural networks},
  author={Geirhos, Robert and Jacobsen, J{\"o}rn-Henrik and Michaelis, Claudio and Zemel, Richard and Brendel, Wieland and Bethge, Matthias and Wichmann, Felix A},
  journal={Nature Machine Intelligence},
  volume={2},
  number={11},
  pages={665--673},
  year={2020},
  publisher={Nature Publishing Group}
}

@article{shah2020pitfalls,
  title={The pitfalls of simplicity bias in neural networks},
  author={Shah, Harshay and Tamuly, Kaustav and Raghunathan, Aditi and Jain, Prateek and Netrapalli, Praneeth},
  journal={Advances in Neural Information Processing Systems},
  volume={33},
  pages={9573--9585},
  year={2020}
}

@article{bellinger2020remix,
  title={Remix: Calibrated resampling for class imbalance in deep learning},
  author={Bellinger, Colin and Corizzo, Roberto and Japkowicz, Nathalie},
  journal={arXiv preprint arXiv:2012.02312},
  year={2020}
}

@inproceedings{recht2019imagenet,
  title={Do imagenet classifiers generalize to imagenet?},
  author={Recht, Benjamin and Roelofs, Rebecca and Schmidt, Ludwig and Shankar, Vaishaal},
  booktitle={International conference on machine learning},
  pages={5389--5400},
  year={2019},
  organization={PMLR}
}

@article{neyshabur2017exploring,
  title={Exploring generalization in deep learning},
  author={Neyshabur, Behnam and Bhojanapalli, Srinadh and McAllester, David and Srebro, Nati},
  journal={Advances in neural information processing systems},
  volume={30},
  year={2017}
}

@article{dablain2022understanding,
  title={Understanding CNN Fragility When Learning With Imbalanced Data},
  author={Dablain, Damien and Jacobson, Kristen N and Bellinger, Colin and Roberts, Mark and Chawla, Nitesh},
  journal={Machine Learning},
  year={April 11, 2023}
}

@article{krawczyk2016learning,
  title={Learning from imbalanced data: open challenges and future directions},
  author={Krawczyk, Bartosz},
  journal={Progress in Artificial Intelligence},
  volume={5},
  number={4},
  pages={221--232},
  year={2016},
  publisher={Springer}
}

@article{he2009learning,
  title={Learning from imbalanced data},
  author={He, Haibo and Garcia, Edwardo A},
  journal={IEEE Transactions on knowledge and data engineering},
  volume={21},
  number={9},
  pages={1263--1284},
  year={2009},
  publisher={Ieee}
}

@article{japkowicz2002class,
  title={The class imbalance problem: A systematic study},
  author={Japkowicz, Nathalie and Stephen, Shaju},
  journal={Intelligent data analysis},
  volume={6},
  number={5},
  pages={429--449},
  year={2002},
  publisher={IOS Press}
}

@article{neyshabur2014search,
  title={In search of the real inductive bias: On the role of implicit regularization in deep learning},
  author={Neyshabur, Behnam and Tomioka, Ryota and Srebro, Nathan},
  journal={arXiv preprint arXiv:1412.6614},
  year={2014}
}

@article{batista2004study,
  title={A study of the behavior of several methods for balancing machine learning training data},
  author={Batista, Gustavo EAPA and Prati, Ronaldo C and Monard, Maria Carolina},
  journal={ACM SIGKDD explorations newsletter},
  volume={6},
  number={1},
  pages={20--29},
  year={2004},
  publisher={ACM New York, NY, USA}
}

@inproceedings{denil2010overlap,
  title={Overlap versus imbalance},
  author={Denil, Misha and Trappenberg, Thomas},
  booktitle={Canadian conference on artificial intelligence},
  pages={220--231},
  year={2010},
  organization={Springer}
}

@inproceedings{prati2004class,
  title={Class imbalances versus class overlapping: an analysis of a learning system behavior},
  author={Prati, Ronaldo C and Batista, Gustavo EAPA and Monard, Maria Carolina},
  booktitle={Mexican international conference on artificial intelligence},
  pages={312--321},
  year={2004},
  organization={Springer}
}

@inproceedings{garcia2007empirical,
  title={An empirical study of the behavior of classifiers on imbalanced and overlapped data sets},
  author={Garc{\'\i}a, Vicente and S{\'a}nchez, Jose and Mollineda, Ramon},
  booktitle={Iberoamerican congress on pattern recognition},
  pages={397--406},
  year={2007},
  organization={Springer}
}

@article{jo2004class,
  title={Class imbalances versus small disjuncts},
  author={Jo, Taeho and Japkowicz, Nathalie},
  journal={ACM Sigkdd Explorations Newsletter},
  volume={6},
  number={1},
  pages={40--49},
  year={2004},
  publisher={ACM New York, NY, USA}
}

@article{weiss2004mining,
  title={Mining with rarity: a unifying framework},
  author={Weiss, Gary M},
  journal={ACM Sigkdd Explorations Newsletter},
  volume={6},
  number={1},
  pages={7--19},
  year={2004},
  publisher={ACM New York, NY, USA}
}

@article{dablain2023towards,
  title={Towards Understanding How Data Augmentation Works with Imbalanced Data},
  author={Dablain, Damien A and Chawla, Nitesh V},
  journal={arXiv preprint arXiv:2304.05895},
  year={2023}
}

@article{cao2019learning,
  title={Learning imbalanced datasets with label-distribution-aware margin loss},
  author={Cao, Kaidi and Wei, Colin and Gaidon, Adrien and Arechiga, Nikos and Ma, Tengyu},
  journal={arXiv preprint arXiv:1906.07413},
  year={2019}
}

@article{dablain2022deepsmote,
  title={DeepSMOTE: Fusing deep learning and SMOTE for imbalanced data},
  author={Dablain, Damien and Krawczyk, Bartosz and Chawla, Nitesh V},
  journal={IEEE Transactions on Neural Networks and Learning Systems},
  year={2022},
  publisher={IEEE}
}

@article{scikit-learn,
 title={Scikit-learn: Machine Learning in {P}ython},
 author={Pedregosa, F. and Varoquaux, G. and Gramfort, A. and Michel, V.
         and Thirion, B. and Grisel, O. and Blondel, M. and Prettenhofer, P.
         and Weiss, R. and Dubourg, V. and Vanderplas, J. and Passos, A. and
         Cournapeau, D. and Brucher, M. and Perrot, M. and Duchesnay, E.},
 journal={Journal of Machine Learning Research},
 volume={12},
 pages={2825--2830},
 year={2011}
}

@article{paszke2017automatic,
  title={Automatic differentiation in pytorch},
  author={Paszke, Adam and Gross, Sam and Chintala, Soumith and Chanan, Gregory and Yang, Edward and DeVito, Zachary and Lin, Zeming and Desmaison, Alban and Antiga, Luca and Lerer, Adam},
  year={2017}
}

\end{document}